\documentclass{article}
\usepackage[nonatbib, final]{nips_2017}
\usepackage[utf8]{inputenc} 
\usepackage[T1]{fontenc}    
\usepackage{hyperref}       
\usepackage{url}            
\usepackage{booktabs}       
\usepackage{amsfonts}       
\usepackage{nicefrac}       
\usepackage{microtype}      
\usepackage{graphicx}
\usepackage{amsmath}
\usepackage{bm}
\usepackage{algorithmicx}
\usepackage{algorithm}
\usepackage{algpseudocode}
\usepackage[table]{xcolor}

\title{AlphaZero Gomoku}

\author{%
  Wen Liang \\
  University of California, San Diego \\
  La Jolla, CA 92093 \\
  \texttt{wel245@eng.ucsd.edu} \\
  \And
  Chao Yu \\
  University of California, San Diego \\
  La Jolla, CA 92093 \\
  \texttt{chy018@eng.ucsd.edu} \\
  \And
  Brian Whiteaker \\
  University of California, San Diego \\
  La Jolla, CA 92093 \\
  \texttt{bwhiteak@ucsd.edu} \\
  \And
  Inyoung Huh \\
  University of California, San Diego \\
  La Jolla, CA 92093 \\
  \texttt{i1huh@ucsd.edu} \\
  \And
  Hua Shao \\
  University of California, San Diego \\
  La Jolla, CA 92093 \\
  \texttt{h5shao@eng.ucsd.edu} \\
  \And
  Youzhi Liang \\
  Stanford University \\
  Stanford, CA 94305 \\
  \texttt{youzhil@stanford.edu} \\
}

\begin{document}

\maketitle

\begin{abstract}


In the past few years, AlphaZero's exceptional capability in mastering intricate board games has garnered considerable interest. Initially designed for the game of Go, this revolutionary algorithm merges deep learning techniques with the Monte Carlo tree search (MCTS) to surpass earlier top-tier methods. In our study, we broaden the use of AlphaZero to Gomoku, an age-old tactical board game also referred to as "Five in a Row." Intriguingly, Gomoku has innate challenges due to a bias towards the initial player, who has a theoretical advantage. To add value, we strive for a balanced game-play. Our tests demonstrate AlphaZero's versatility in adapting to games other than Go. MCTS has become a predominant algorithm for decision processes in intricate scenarios, especially board games. MCTS creates a search tree by examining potential future actions and uses random sampling to predict possible results. By leveraging the best of both worlds, the AlphaZero technique fuses deep learning from Reinforcement Learning with the balancing act of MCTS, establishing a fresh standard in game-playing AI. Its triumph is notably evident in board games such as Go, chess, and shogi.

\end{abstract}

\section{Introduction}

Reinforcement learning (RL) is a pivotal and rapidly advancing domain within contemporary artificial intelligence research. It offers a distinctive framework wherein agents progressively improve their performance not through explicit instruction but through continual interaction with their surroundings~\cite{kaelbling1996reinforcement, li2017deep}. As these agents take actions in a given environment, they are provided feedback in the form of either rewards for desirable actions or penalties for undesirable ones~\cite{wiering2012reinforcement}. This trial-and-error mechanism aids the agents in understanding the consequences of their actions and refining their strategies accordingly. The primary goal of RL is to determine an optimal strategy, often termed as a "policy", which instructs the agent on the best possible action to take in any given situation. The optimal policy is one that, when followed, will lead to the maximization of the cumulative rewards over a period, ensuring that the agent's actions result in the most favorable outcomes in its environment. Board games, with their intricate complexities and well-defined reward structures, make a fitting domain for RL, offering a perspective ripe for academic inquiry. MCTS (Monte Carlo tree search)~\cite{mtcs} has emerged as a leading algorithm for decision-making in these complex environments. Recently, deep learning catalyzed groundbreaking developments in diverse research domains, from computer vision and natural language processing to state-of-the-art recommender systems~\cite{miamimx, transformer, recommender}. It constructs a search tree by exploring possible future moves, using statistical sampling to evaluate the potential outcomes. It constructs a search tree by exploring possible future moves, using statistical sampling to evaluate the potential outcomes. Bridging the gap between RL and MCTS, the original AlphaGo~\cite{alphago} algorithm showcased a fusion of deep learning and tree search techniques, revolutionizing the game-playing AI landscape. This groundbreaking approach further evolved with the introduction of AlphaZero~\cite{alphazero}, which uses zero human knowledge and experience in this game and removes the supervised learning stage. It allows the algorithm to master the game with self-learning only.

Gomoku, often referred to as "Five in a Row," is usually played on a 15x15 grid (though variations can feature larger grids). The game's objective is straightforward yet captivating: two players, typically designated as black and white, take turns placing stones on the board with the aim to align five of their own stones consecutively in a vertical, horizontal, or diagonal line. The game's seemingly simple rules mask a depth of strategy. Early moves tend to be concentrated around the center of the board, providing players with maximal opportunities to expand and form their sequences. As the game progresses, the board transforms into a complex battleground of potential sequences, blocked attempts, and intricate traps. The nature of the game allows for both defensive and offensive tactics. A player might focus on preventing their opponent from completing a sequence, or strategically placing their stones to create multiple potential winning avenues simultaneously. Its straightforward rules combined with its profound complexity make it an ideal candidate for studying artificial intelligence's prowess and potential in mastering classic board games. In recent decades, there have been concerted efforts to solve Gomoku using computational methods. One notable attempt was by Allis~\cite{allis}, who employed proof-number search algorithms to analyze specific game positions and paths, making significant strides in understanding the game's complexities. Another significant contribution came from Chen~\cite{chen}, who utilized pattern recognition and threat-space search techniques to advance AI capabilities in Gomoku, offering a fresh perspective on potential winning strategies.

Driven by the recent monumental strides in board game artificial intelligence, especially the unparalleled triumphs of the AlphaZero algorithm, we were compelled to believe that harnessing this cutting-edge approach for the game of Gomoku was not only feasible but imperative. In embarking on this ambitious journey, our contributions to the Gomoku AI research landscape manifest in two significant dimensions:

\begin{enumerate}
    \item We generalized the AlphaZero approach for the Gomoku game, achieving impressive results. Initiating from a state of random play, and without any domain knowledge apart from the game rules, our model swiftly learned a winning strategy on a 6x6 table after just a few hours of training on an economical GPU.
    

    \item We embarked on an extensive research endeavor, wherein we juxtaposed the efficacy of our refined AlphaZero methodology against a conventional method that exclusively leverages the Monte Carlo tree search (MCTS). Our aim was to critically assess how these two distinct techniques fare in terms of both efficiency and effectiveness under comparable conditions, to shed light on their relative strengths and potential areas of improvement.

\end{enumerate}

\section{Method}

\subsection{Value and Policy Network}

Deep neural network in AlphaZero~\cite{alphazero} and AlphaGo~\cite{alphago} often employs two primary neural networks: the Value Network (\( V \)) and the Policy Network (\( \pi \)). 

\begin{itemize}
    \item \textbf{Value Network (\( V \))}: This network estimates the value of a given state, i.e., the expected outcome from that state. Formally, for a given state \( s \), \( V(s) \) predicts the expected outcome, with values close to +1 indicating favorable outcomes for the player and values close to -1 indicating unfavorable outcomes.
    \begin{equation}
    V(s) \approx \mathbb{E}[r|s]
    \end{equation}
    where \( \mathbb{E} \) is the expectation and \( r \) is the eventual reward.

    \item \textbf{Policy Network (\( \pi \))}: This network provides a probability distribution over all possible moves from a given state. For a state \( s \) and an action \( a \), \( \pi(a|s) \) represents the probability of taking action \( a \) as a highly optimized game player.
    \begin{equation}
    \pi(a|s) = P(a \text{ is the best move | } s)
    \end{equation}
\end{itemize}

\begin{figure} [H]
	\centering
    \includegraphics[width=0.6\textwidth]{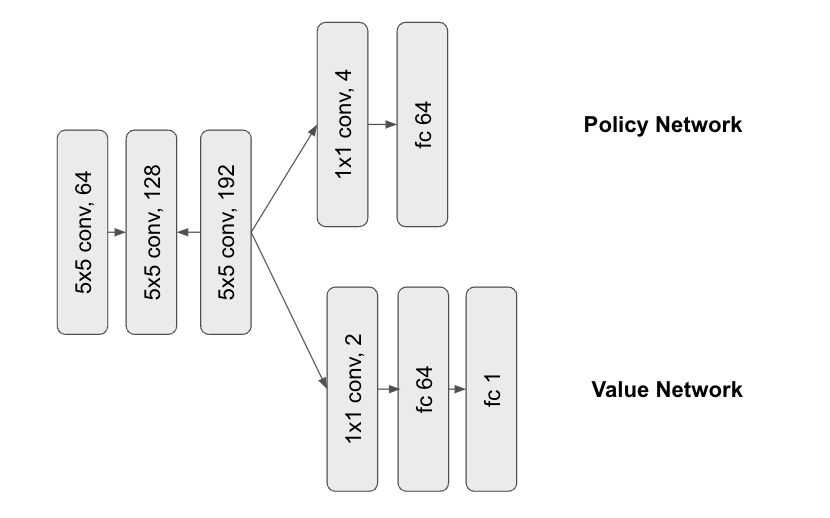}
    \caption{Value and Policy Networks}
    \label{fig:network}
\end{figure}

The neural network structure we use is shown in Figure~\ref{fig:network}.

\subsection{Monte Carlo Tree Search (MCTS)}




Monte Carlo Tree Search (MCTS) stands out as a revolutionary algorithm, reshaping decision-making processes in intricate environments through the methodical construction of a search tree. At its core, the algorithm meticulously evaluates prospective game moves, striking a harmonious equilibrium between the dual tenets of exploration (unearthing new moves) and exploitation (leveraging known advantageous moves). The integration of Policy and Value networks into this framework bestows it with unparalleled depth and precision:

\begin{itemize}
\item The Policy Network, through its discerning output, serves as the beacon guiding the expansion of the search tree. Instead of branching out indiscriminately, it casts the spotlight on moves radiating promise and potential, ensuring that the exploration process remains strategic and focused.
\item On the other hand, the Value Network steps in as an adept evaluator, meticulously scrutinizing leaf nodes within the tree. This network diminishes the traditional reliance on random rollouts for evaluation, infusing the process with a heightened level of precision. This capability not only speeds up the evaluation but also endows it with a more profound insight into the game's dynamics.
\end{itemize}

In essence, the Policy Network acts as a compass, navigating the vast possibilities in the MCTS landscape and directing it towards potentially rewarding paths. Simultaneously, the Value Network functions as an astute analyst, swiftly gauging the potential outcomes of different game scenarios. Their synergistic interplay ensures that the search process within MCTS remains both streamlined and enriched, leading to decisions that are both efficient and strategically sound.

\subsection{Environment and Supervised Learning}

\begin{figure} [H]
	\centering
    \includegraphics[width=0.6\textwidth]{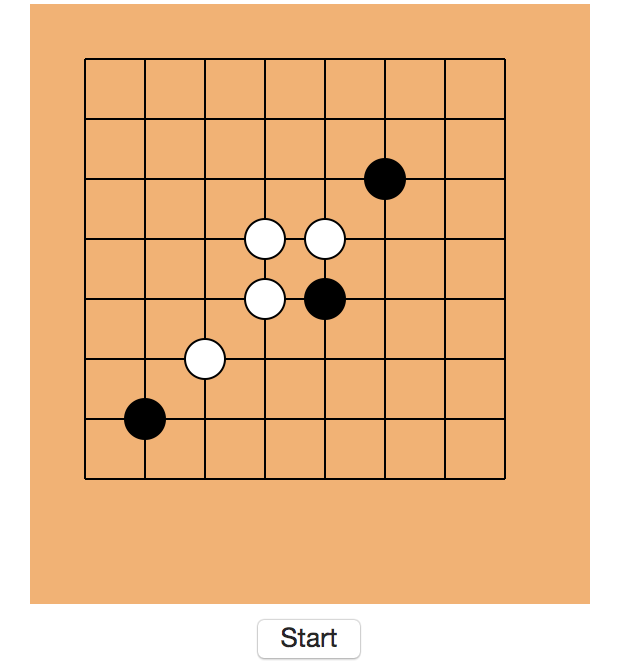}
    \caption{Gomoku game board from our implementation}
    \label{fig:board}
\end{figure}

Within the realm of reinforcement learning, our agent actively interacts with a specially-designed Gomoku gaming environment, drawing feedback in the form of rewards or penalties based on its moves. As depicted in Figure\ref{fig:board}, we meticulously implemented a Gomoku game board that closely mirrors traditional gameplay. Given the computational overhead associated with larger boards, we strategically focused our experimental investigations on boards sized $6\times 6$, targeting a 4-in-a-row win condition, and $8\times8$, targeting the standard 5-in-a-row. To provide a comprehensive representation of each game state, we innovatively devised four distinct binary feature matrices. These matrices encapsulate essential game facets, including the current player's move, the adversary's move, the most recent move, and the initiating player. Notably, these matrices not only serve as a holistic game state representation but also act as pivotal input layers for our deep learning neural network. In terms of game mechanics, we faithfully incorporated Gomoku's conventional victory criteria. Moves are delineated at board intersections, eschewing placement within board squares. Traditionally, the white player initiates the game, with both players alternating their moves in succession until a conclusive game outcome is achieved. The essence of the game revolves around players positioning a stone of their chosen color on an unoccupied intersection. Triumph is heralded by the first player to strategically place five of their stones consecutively, irrespective of orientation - horizontal, vertical, or diagonal. In the rare scenario where the board reaches saturation without either contender achieving the coveted five-in-a-row, the game is ceremoniously declared a stalemate. While Gomoku might superficially seem straightforward, it belies a profound strategic depth, characterized by its myriad winning patterns and tactical nuances.


Gomoku's strategic intricacies are underscored by the delicate balance and importance of specific board configurations, notably the 'threes' and 'fours'. These patterns play pivotal roles in dictating the pace and outcome of a match. When leveraged effectively, they can swiftly shift the advantage to one player, often pushing the opponent into a corner from which recovery becomes arduous.

Diving into the nuances, the 'four' configuration is a fascinating tactical alignment where four stones of identical color stand in unison, beckoning a potential game-winning fifth stone in the subsequent move. The looming threat of this alignment is palpable, sending clear signals of an impending victory. Recognizing this, an opponent is thrust into a defensive mode, compelled to respond instantly. Failure to address this sequence by obstructing the alignment invariably results in a loss, testament to its lethal efficacy.

\begin{figure}[H]
\centering
  \centering
  \includegraphics[width=.6\linewidth]{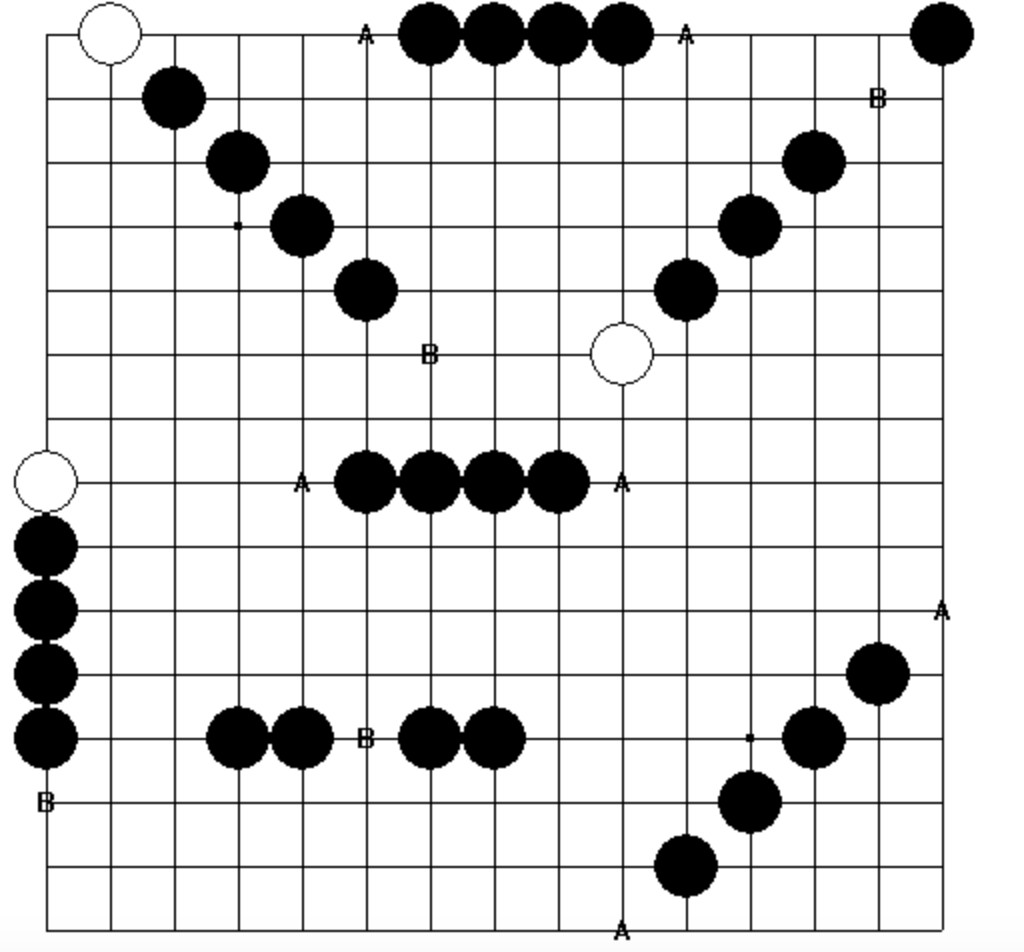}
  \caption{'Four' winning pattern}
  \label{fig:sub1}
\end{figure}

\begin{figure}
  \centering
  \includegraphics[width=.6\linewidth]{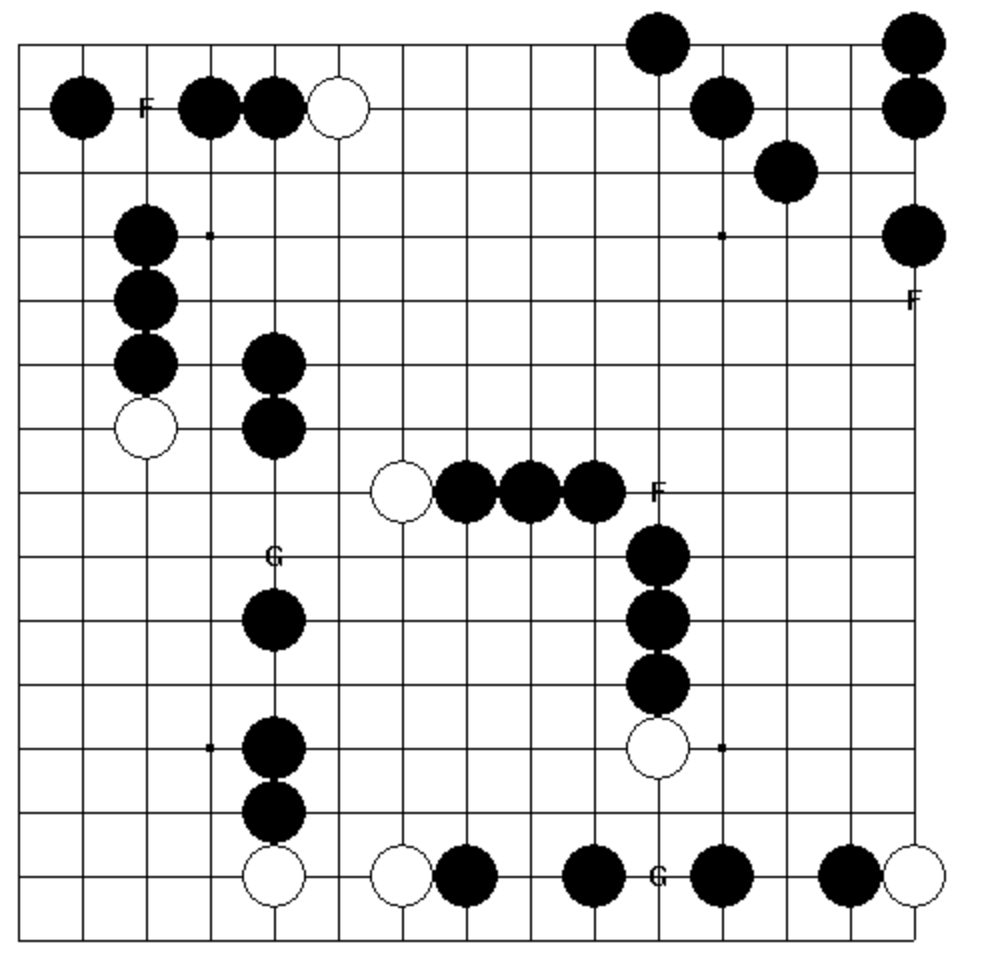}
  \caption{'Fork' winning pattern}
  \label{fig:sub2}
\end{figure}

Equally compelling, yet distinct in its strategic implications, is the 'fork' configuration. In this maneuver, a player crafts a masterstroke with a single move, spawning two formidable attack sequences in tandem. The duality of the threat is what sets the 'fork' apart: it presents a dual quandary that the opponent must navigate. The challenge is steep; it's nearly impossible to stymie both threats concurrently. Thus, successfully engineering a fork is often tantamount to clinching the game, rendering the adversary powerless in the face of this double-edged assault. To truly appreciate the visual elegance and tactical profundity of these configurations, one can refer to Figure~\ref{fig:sub1} and Figure~\ref{fig:sub2}. These illustrations vividly capture the essence of 'threes', 'fours', and the enigmatic 'fork', underscoring their pivotal roles in the beautiful complexity of Gomoku.

\section{Result}


Our experimentation painted an optimistic picture when applying the AlphaZero methodology to the Gomoku game. Significantly, our rendition not only succeeded but boasted an impeccable 100\% victory rate as the initiating player during self-play assessments. Moreover, as the succeeding player, the algorithm manifested a keen aptitude for defense, coupled with a proactive stance towards identifying and capitalizing on counterattack chances. A detailed exemplification of this nuanced behavior is cataloged in Appendix I.

\begin{figure}
  \centering
  \includegraphics[width=.6\linewidth]{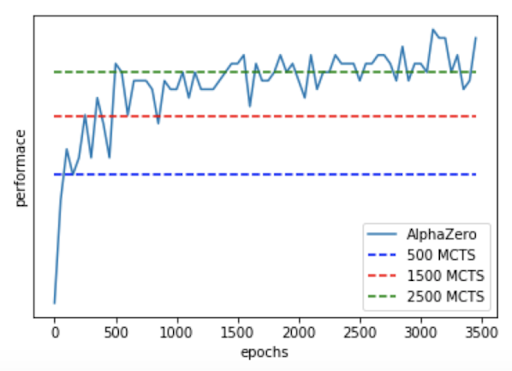}
  \caption{Compare AlphaZero with MTCS}
  \label{fig:compare}
\end{figure}

In our research, we undertook an in-depth comparative study, juxtaposing the performance of the AlphaZero approach against the traditional Monte Carlo tree search (MCTS) method. To furnish a comprehensive perspective, we analyzed a spectrum of iterations, spanning from 500 to 2500 in number. The ensuing patterns and performance distinctions are graphically represented in Figure~\ref{fig:compare}. The empirical data unambiguously accentuates AlphaZero's superior efficacy, as it continually eclipses the performance benchmarks set by the MCTS method.

\section{Conclusion}


We meticulously crafted a simulation environment tailored for the game of Gomoku and, within this context, developed a specialized agent. By adopting and integrating the AlphaZero methodology into our platform, we were able to achieve not only functional outcomes but also results that exceeded our initial expectations. These findings underscore the profound efficacy of the AlphaZero technique in mastering intricate board games, such as Gomoku, demonstrating its versatility and robustness in diverse gaming scenarios.

\bibliographystyle{unsrt}
\bibliography{reference}

\section*{Appendix I}

This section showcases a set of self-play match results on a \(6 \times 6\) board, where a sequence of four consecutive pieces leads to a win. In the presented board, the 'x' pieces represent the first player, while the 'o' pieces symbolize the second player. The darker shades on the board correspond to positions where the policy network predicted a stronger advantage.

\begin{figure} [H]
	\centering
    \includegraphics[width=0.6\textwidth]{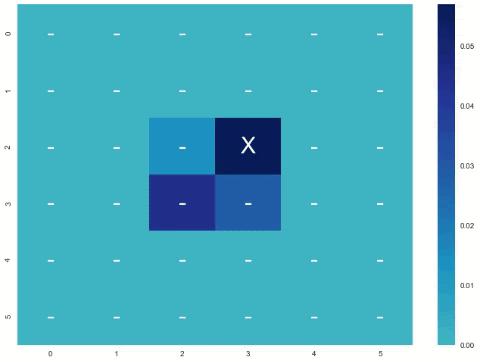}
    \caption{Example step 1}
    \label{fig:step1}
\end{figure}

\begin{figure} [H]
	\centering
    \includegraphics[width=0.6\textwidth]{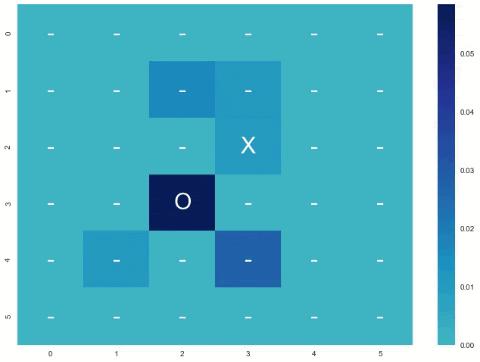}
    \caption{Example step 2}
    \label{fig:step2}
\end{figure}

\begin{figure} [H]
	\centering
    \includegraphics[width=0.6\textwidth]{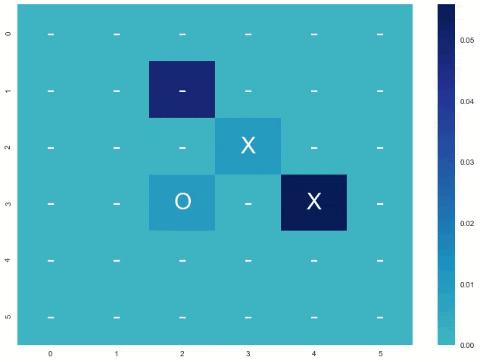}
    \caption{Example step 3}
    \label{fig:step3}
\end{figure}

\begin{figure} [H]
	\centering
    \includegraphics[width=0.6\textwidth]{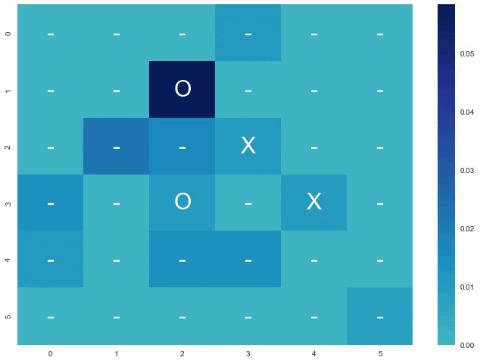}
    \caption{Example step 4}
    \label{fig:step4}
\end{figure}

\begin{figure} [H]
	\centering
    \includegraphics[width=0.6\textwidth]{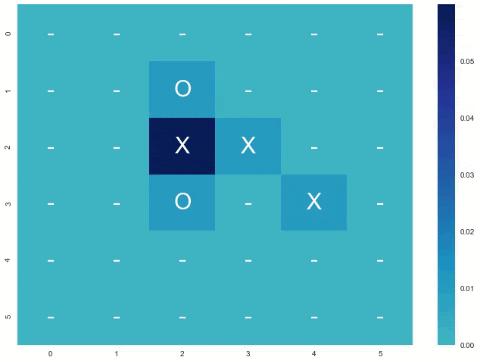}
    \caption{Example step 5}
    \label{fig:step5}
\end{figure}

\begin{figure} [H]
	\centering
    \includegraphics[width=0.6\textwidth]{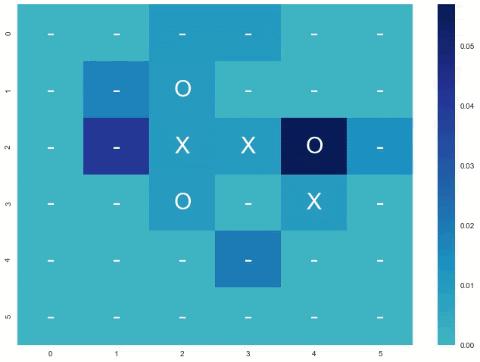}
    \caption{Example step 6}
    \label{fig:step6}
\end{figure}

\begin{figure} [H]
	\centering
    \includegraphics[width=0.6\textwidth]{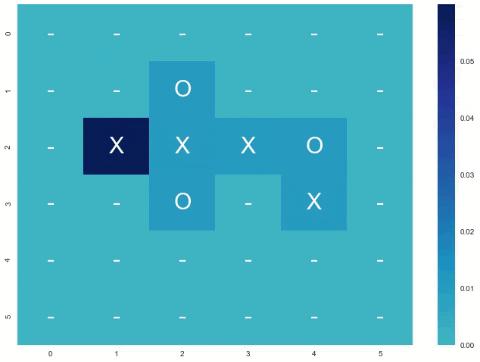}
    \caption{Example step 7}
    \label{fig:step7}
\end{figure}

\begin{figure} [H]
	\centering
    \includegraphics[width=0.6\textwidth]{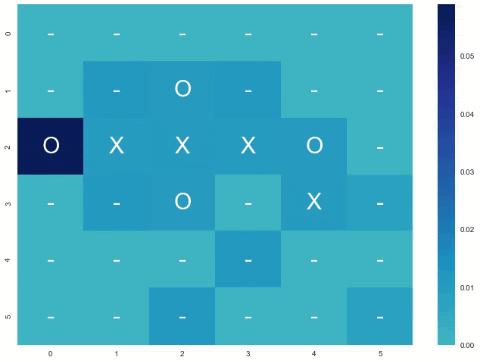}
    \caption{Example step 8}
    \label{fig:step8}
\end{figure}

\begin{figure} [H]
	\centering
    \includegraphics[width=0.6\textwidth]{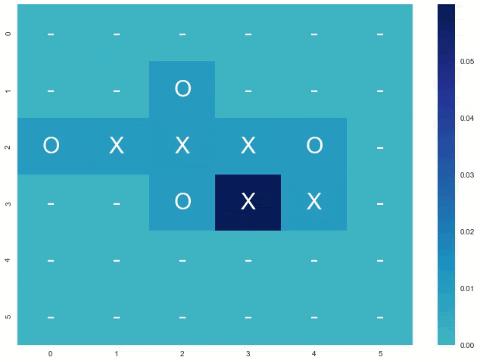}
    \caption{Example step 9}
    \label{fig:step9}
\end{figure}

\begin{figure} [H]
	\centering
    \includegraphics[width=0.6\textwidth]{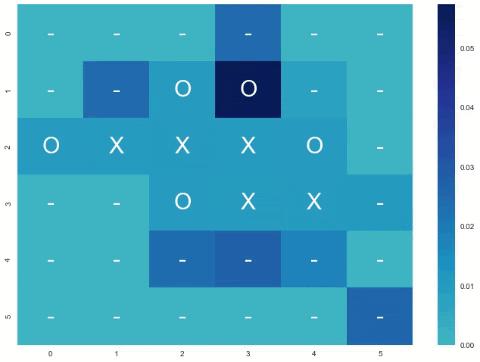}
    \caption{Example step 10}
    \label{fig:step10}
\end{figure}

\begin{figure} [H]
	\centering
    \includegraphics[width=0.6\textwidth]{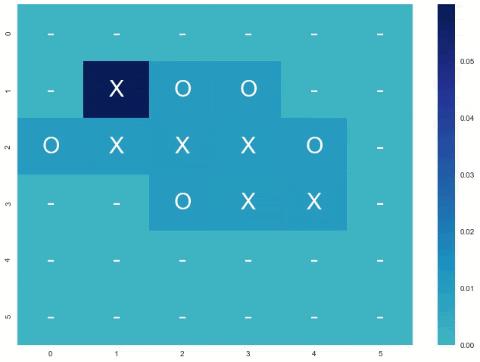}
    \caption{Example step 11}
    \label{fig:step11}
\end{figure}

\begin{figure} [H]
	\centering
    \includegraphics[width=0.6\textwidth]{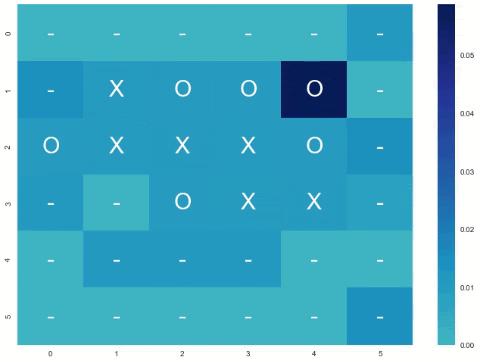}
    \caption{Example step 12}
    \label{fig:step12}
\end{figure}

\begin{figure} [H]
	\centering
    \includegraphics[width=0.6\textwidth]{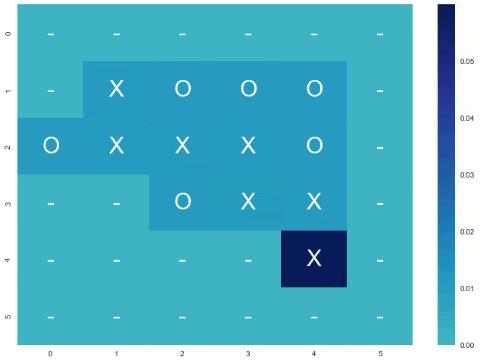}
    \caption{Example step 13}
    \label{fig:step13}
\end{figure}

\end{document}